%% file: main.tex
\documentclass[10pt,twocolumn,letterpaper]{article}

\usepackage{cvpr}              

\input{preamble}

\definecolor{cvprblue}{rgb}{0.21,0.49,0.74}
\usepackage[pagebackref,breaklinks,colorlinks,citecolor=cvprblue]{hyperref}

\def\paperID{3} 
\def\confName{CVPR}
\def\confYear{2024}

\title{V-VIPE: Variational View Invariant Pose Embedding}

\author{Mara Levy\\
University of Maryland, College Park\\
{\tt\small mlevy@umd.edu}
\and
Abhinav Shrivastava\\
University of Maryland, College Park\\
{\tt\small abhinav@cs.umd.edu}
}

\begin{document}
\maketitle
\input{sec/0_abstract}    
\input{sec/1_intro}
\input{sec/2_related_work}
\input{sec/3_method}
\input{sec/4_results}
\input{sec/5_conclusion}
{
    \small
    \bibliographystyle{ieeenat_fullname}
    \bibliography{main}
}


\end{document}


\maketitle
\input{sec/X_suppl}
{
    \small
    \bibliographystyle{ieeenat_fullname}
    \bibliography{main}
}


%% file: preamble.tex
\usepackage[dvipsnames]{xcolor}


%% file: sec/0_abstract.tex
\begin{abstract}
Learning to represent three dimensional (3D) human pose given a two dimensional (2D) image of a person, is a challenging problem. In order to make the problem less ambiguous it has become common practice to estimate 3D pose in the camera coordinate space. However, this makes the task of comparing two 3D poses difficult. In this paper, we address this challenge by separating the problem of estimating 3D pose from 2D images into two steps. We use a variational autoencoder (VAE) to find an embedding that represents 3D poses in canonical coordinate space. We refer to this embedding as variational view-invariant pose embedding (\textbf{V-VIPE}). Using V-VIPE we can encode 2D and 3D poses and use the embedding for downstream tasks, like retrieval and classification. We can estimate 3D poses from these embeddings using the decoder as well as generate unseen 3D poses. The variability of our encoding allows it to generalize well to unseen camera views when mapping from 2D space. To the best of our knowledge, V-VIPE is the only representation to offer this diversity of applications. Code and more information can be found at https://v-vipe.github.io/.
\end{abstract}

%% file: sec/1_intro.tex
\section{Introduction}
\label{sec:intro}
Learning to represent three dimensional (3D) human pose given a two dimensional (2D) image of a person, is a challenging problem with several important downstream applications such as teaching a person to mimic a video, action recognition and imitation learning for robotics. The key challenge arises from the fact that different camera viewpoints observing the same 3D pose lead to very different projections in a 2D image. The common practice is to circumvent this challenge by estimating 3D pose in the camera coordinate space~\cite{DBLP:journals/corr/ZhouH0XW17, DBLP:journals/corr/MartinezHRL17, DBLP:journals/corr/abs-1711-08585}. However, this leads to differences in scale and rotation between the estimated 3D representations from images of the same 3D pose from different camera viewpoints. Without the knowledge of camera parameters, it is not possible to establish correspondence between these 3D representations.  This is important as we move towards environments where we have very little control over the camera viewpoint, such as photos taken with a phone or AR glasses. In such scenarios, we can make very few assumptions about the camera space.

\begin{figure}[t]
   \includegraphics[ width=\linewidth]{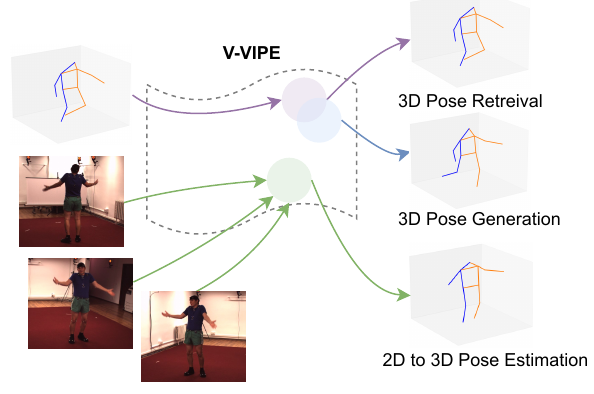}
   \caption{The several functions V-VIPE is capable of. The purple path represents 3D pose retrieval. The blue path represents generation by adding noise to the purple path. The result is a variation of the original pose. The green path shows 2D to 3D pose estimation from several viewpoints.}
   \vspace{-0.2in}
   \label{fig:teaser}
\end{figure}

In this paper, we address this challenge by separating the problem of estimating 3D pose from 2D images into two steps. First, we learn an embedding to represent 3D poses in canonical coordinate space. Next, we learn to encode 2D poses, from different camera viewpoints, to the embedding from the first step. This leads to a canonical 3D pose embedding that is invariant to camera viewpoints. This view-invariant pose embedding is highly flexible, allowing us to do 3D pose retrieval, 3D pose generation, and most importantly, estimating consistent 3D pose from different 2D viewpoints~\ref{fig:teaser}.

In our approach we use a variational autoencoder (VAE) to learn an embedding for 3D human poses. This VAE is trained to reconstruct 3D poses and has two key benefits: (a) we can leverage loss functions to ensure similar 3D poses are close in the embedding space, and (b) we learn embeddings that can generalize better to unseen 3D poses due to the variational training paradigm. Next, we learn a mapping from 2D poses (either ground-truth or estimated using off-the-shelf detectors) to this 3D pose embedding space by training a 2D pose encoder that estimates the 3D pose embedding. This embedding is used as input to the pre-trained decoder from the VAE to estimate the corresponding 3D pose, thus leading to ``lifting'' the 2D pose from different camera viewpoints to 3D~\cite{DBLP:journals/corr/ParkHK16, Katircioglu2018LearningLR}. We refer to embedding as variatonal view-invariant pose embedding (\textbf{V-VIPE}).

Our proposed V-VIPE is highly flexible and generalizable. We can encode 3D poses and use the embedding for downstream tasks, like retrieval and classification. We can also map 2D poses from unseen camera viewpoints to this embedding. We can estimate 3D poses from these embeddings using the decoder. Finally, we can generate unseen 3D poses.  To the best of our knowledge, V-VIPE is the only representation to offer this diversity of applications.\looseness=-1

We perform an extensive experimental evaluation over two datasets: Human 3.6M~\cite{h36m_pami} and MPI-3DHP~\cite{mono-3dhp2017}. We show quantitative results on 2D to 3D pose retrieval and qualitative results on 3D pose generation and 2D to 3D pose estimation. We show that V-VIPE performs $1\%$ better than other embedding methods on seen camera viewpoints and about $2.5\%$ better for unseen camera viewpoints. In addition, we show generalization of our approach by training on one dataset and testing on the other.

To summarize, our main contributions are as follows:
\begin{itemize}
\setlength\itemsep{0em}
\item We learn a variational view-invariant pose embedding (V-VIPE) by training a VAE to represent 3D poses in canonical coordinate space, which allows it to be camera invariant.
\item We propose a model to map from 2D poses to V-VIPE, which enables us to estimate 3D poses of 2D images. 
Additionally, because V-VIPE is camera invariant, our mapping can generalize to unseen cameras.
\item We also estimate and generate 3D poses using V-VIPE via a decoder that can be used for downstream tasks.
\end{itemize}

In the rest of the paper we expand upon these ideas. We summarize the related works in Section~\ref{sec:related_work}. In Section~\ref{sec:proposed_method}, we describe our proposed method, and Section~\ref{sec:exp_res} provides the experimental evaluations. Section~\ref{sec:ablation} looks at ablations of our method. Finally, Section~\ref{sec:conclusion} derives the conclusions.

%% file: sec/2_related_work.tex
\section{Related Work}
\label{sec:related_work}

\begin{figure*}[t]
  \centering
  \vspace{-0.1in}
   \includegraphics[trim=1cm 95cm 78cm 2cm, width=\linewidth]{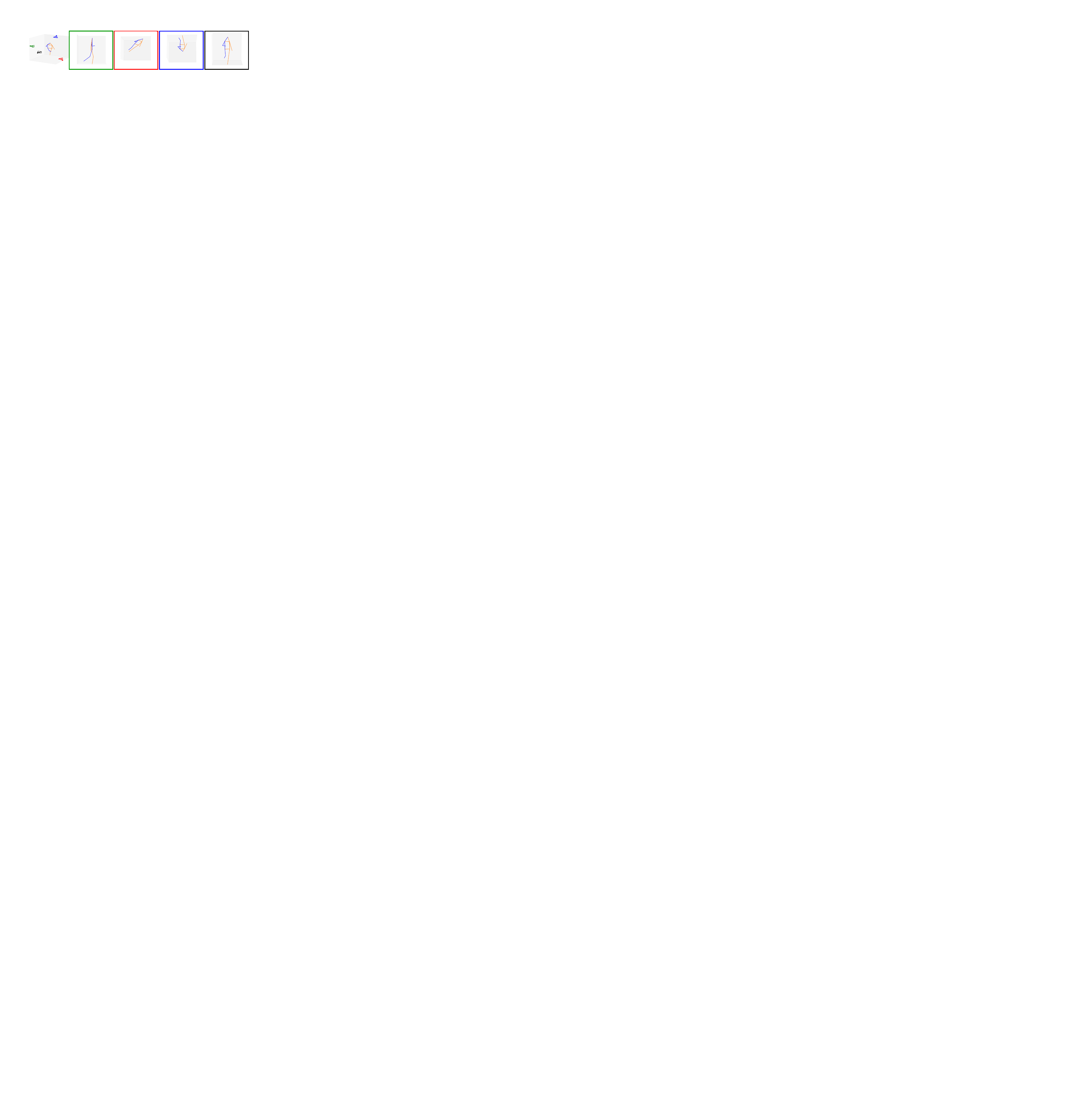}
   \caption{On the left we can see the 3D pose in the original global coordinates with 4 different cameras. The next 4 images are the 3D poses as seen from these 4 cameras.}
   \label{fig:global_rotation}
\end{figure*}

\noindent\textbf{Human Pose Estimation.}
There are two family of approaches for human pose estimation. One is to directly estimate 3D poses from an 2D images~\cite{DBLP:journals/corr/PavlakosZDD16, DBLP:journals/corr/abs-1711-08229}, and the other is to lift pre-detected 2D poses to 3D poses~\cite{DBLP:journals/corr/MartinezHRL17, DBLP:journals/corr/abs-1912-01001, DBLP:journals/corr/abs-2007-09389}. In recent years, state-of-the-art approaches have almost exclusively focused on the lifting strategy. 

Our goal is to specifically find correspondence between 2D poses in images from different camera viewpoints without any knowledge of camera parameters or temporal context.  Recent works have explored how temporal information can improve 3D pose estimation~\cite{DBLP:journals/corr/abs-2103-10455, DBLP:journals/corr/abs-2004-11822}, typically by processing a sequence of images using a transformer~\cite{DBLP:journals/corr/abs-2103-10455}. However, our focus is 3D pose estimation using a single 2D image which is similar to~\cite{DBLP:journals/corr/abs-1901-10841}.

The key distinction between our approach and prior works in estimating 3D poses using 2D images is view-invariant embedding that can be estimated from a monocular viewpoint. Several works have attempted to address view invariant estimation by leveraging many viewpoints~\cite{DBLP:journals/corr/abs-2111-04076, DBLP:journals/corr/abs-2004-02186, DBLP:journals/corr/abs-2110-09554} because it is much easier to place a person in canonical coordinate space when you have access to many views. However, access to multiple viewpoints of the same scene is an unrealistic assumption in the canonical settings. Therefore, these approaches can only be used in environments that have multiple cameras observing the same scene. In contrast, our approach can be applied to any arbitrary 2D image. Some works only use a single viewpoint during inference time, but still require multiple views for each pose during training~\cite{DBLP:journals/corr/abs-1903-02330}. Whereas our method is more flexible and can be trained on any dataset with both 2D and 3D information, even if there is only one camera viewpoint available. Similar to our work,~\cite{DBLP:journals/corr/abs-1901-10841} performs view-invariant pose estimation from one view, but their method requires localized transformations that fundamentally change the 3D pose and must be reversed at the end to get the final pose. Our approach, on the other hand, requires only one global rotation to a canonical camera viewpoint that does not change the integrity of the pose.

\noindent\textbf{3D Pose Generation.} Training a model capable of generating new 3D poses is important for representing unseen data in addition to training data. There are two main types of generators that can be used, Generational Adversarial Networks (GANs) and Variational Auto Encoders (VAEs).

Several works have used GANs~\cite{9257470, DBLP:journals/corr/abs-1901-10841, DBLP:journals/corr/abs-1803-09722} to generate training data for 3D poses. However, they are not well suited for our task which also requires encoding 3D poses in an embedding space. VAEs, on the other hand, are better suited for learning embedding by auto-encoding 3D poses.~\cite{9506722} learns a latent network, where they go directly from 2D to 3D without using the 3D data as input to the model, whereas~\cite{DBLP:journals/corr/abs-1904-01324} learns a latent representation using a variant of VAE and generate 3D poses using 2D pose as a precondition to their decoder. ~\cite{Katircioglu2018LearningLR} employs a basic autoencoder instead of a VAE, which leads to an inconsistent embedding space that is harder to map to 2D inputs.~\cite{DBLP:journals/corr/GirdharFRG16} also learns an autoencoder instead of a VAE, but additionally, they choose to regress on the embedding and perform little normalization prior to training which leads to a poorly regularized output space.

%% file: sec/3_method.tex
\section{Proposed Method}
\label{sec:proposed_method}

Our method consists of three main parts. In \ref{sec:data_processing} we review the input data pre-processing to ensure that the output is independent of camera view. In section \ref{sec:vae}, we describe how we define V-VIPE through a VAE model. In section \ref{sec:2d} we cover how we learn V-VIPE from the detected keypoints. The final model is a network that takes as input a single frame monocular image and estimates a view invariant pose, which can be used to compare any two human poses independent of the context of the original image.

\subsection{Data Processing}
\label{sec:data_processing}

Before we pass any data through our model we perform two key steps. First, we modify the global rotation of the image; second, we scale the keypoints so that the original size does not affect the model.

\noindent\textbf{Global Rotation Realignment.}
Predicting 3D pose in canonical space is extraordinarily difficult as mentioned in ~\cite{DBLP:journals/corr/MartinezHRL17}. We believe this is mostly due to the global rotation\footnote{By global rotation we mean how a human is rotated in relation to the canonical space.} of any 3D pose. Global rotation is hard to estimate due to its ambiguity. We can see in Figure~\ref{fig:global_rotation} that a pose in global space can have a very different appearance in camera space. Without any information, such as a ground truth pose, which we can align the output to or any camera parameters, it would be difficult to determine that any two of these poses are the same. 

\begin{figure}[t]
  \centering
   \includegraphics[trim=8cm 88cm 89cm 5cm, width=.4\linewidth]{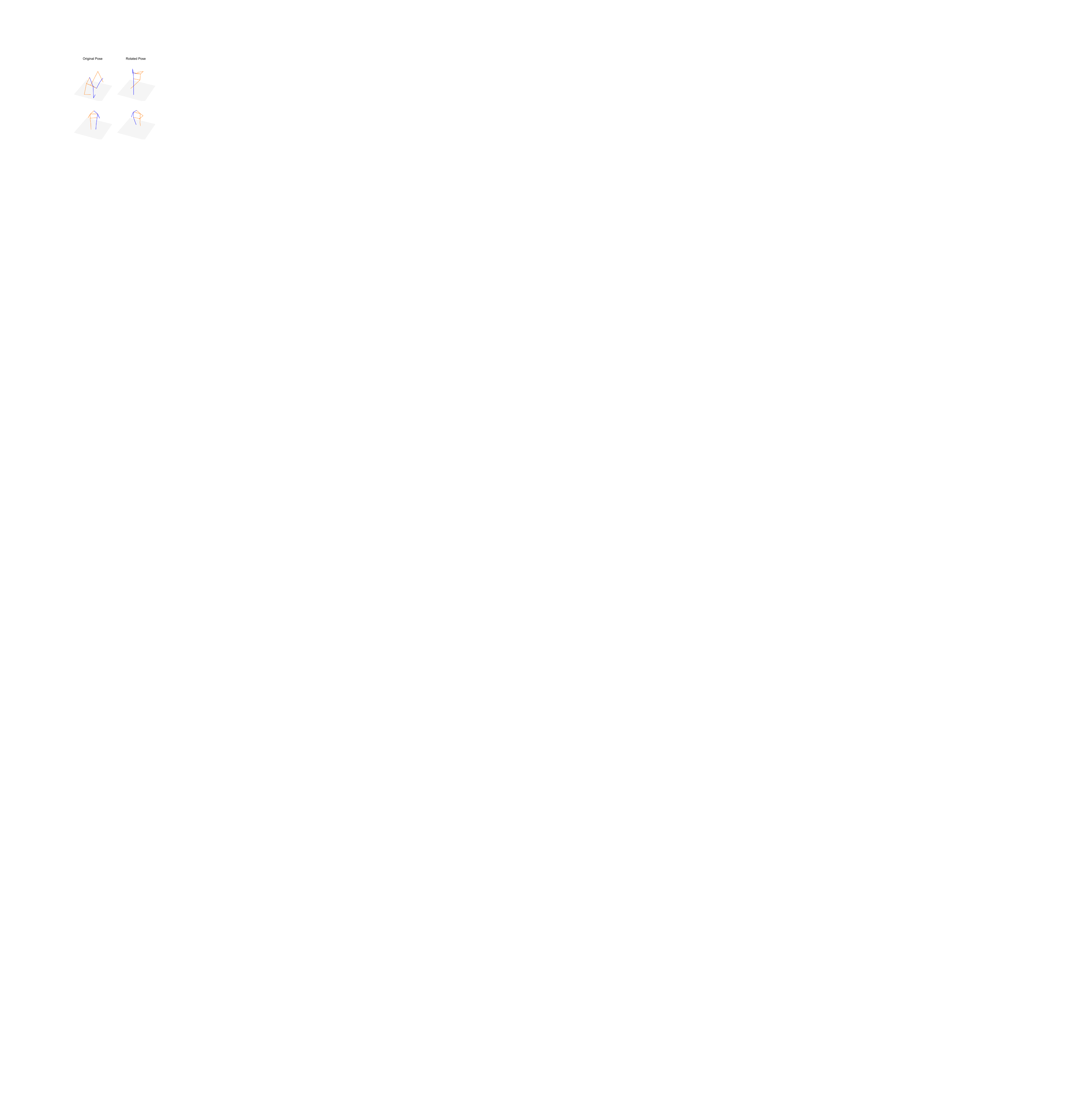}
   \caption{How poses change when we align the points and modify the rotation. On the left is the original pose and on the right is the pose after we have rotated it.}
   \vspace{-0.2in}
   \label{fig:fixed_rotation}
\end{figure}

We argue that global rotation is irrelevant for human pose comparison. Specifically, when we are trying to determine if two poses are the same we do not need to understand how those are oriented in relation to the world they are in. If one pose is facing the x-axis and the other is facing the y-axis, it is still possible that their overall pose is the same. We thus remove rotation dependence by aligning the coordinates of the left hip, right hip and the spine to the same points in every pose of the dataset. This can be visualized in Figure~\ref{fig:fixed_rotation}. In order to achieve such alignment we find the rotation that minimizes the equation: 

\begin{equation}
L(C) = \frac{1}{2}\sum_{i=1}^{n} ||a_i - Cb_i||^2
\label{eq:align}
\end{equation} where $\text{a}_\text{1}, \text{a}_\text{2}, \text{a}_\text{3}$  equal the 3D points representing the left hip, right hip and spine respectively and $\text{b}_\text{1}, \text{b}_\text{2}, \text{b}_\text{3}$ equal $[[0, -1, 0], [0, 1 ,0], [0,0,1]]$. Aligning to these points causes the hips to align to the y axis and the spine to the z axis. We specifically align the hips because they are in a straight line so it is easy to align to one axis and the spine because it is directly above the root and therefore can be easily aligned to a perpendicular axis.  In order to minimize Equation~\ref{eq:align},  we use the Kabsch algorithm~\cite{Kabsch:a15629}. 

\noindent\textbf{Scaling and Pose Normalization.}
\label{sec:normalize}
In this work, we are only concerned with estimating pose such that it is easy to compare how similar two poses are. This is because pose comparison is what is needed for downstream tasks such as action recognition. To account for this, we scale and normalize the input, such that it becomes independent from factors\footnote{Intuitively, two people are capable of being in the same pose no matter their height or weight.} that should not affect the pose similarity estimation.

We use the universal skeleton provided by the dataset to remove the size factor. In this representation all joints are scaled to the same proportions. This makes the size of the 3D output independent of the inputted 2D image or the original 3D pose.

Moreover, to complete the normalization of the data we use a process similar to ~\cite{DBLP:journals/corr/abs-1904-04812} where we center the root joint and scale all of the other joints, accordingly.

\subsection{3D Pose VAE}
\label{sec:vae}

\begin{figure*}[t]
  \centering
  \vspace{-0.3in}
   \includegraphics[trim=0cm 92.5cm 79cm 2cm, width=\linewidth]{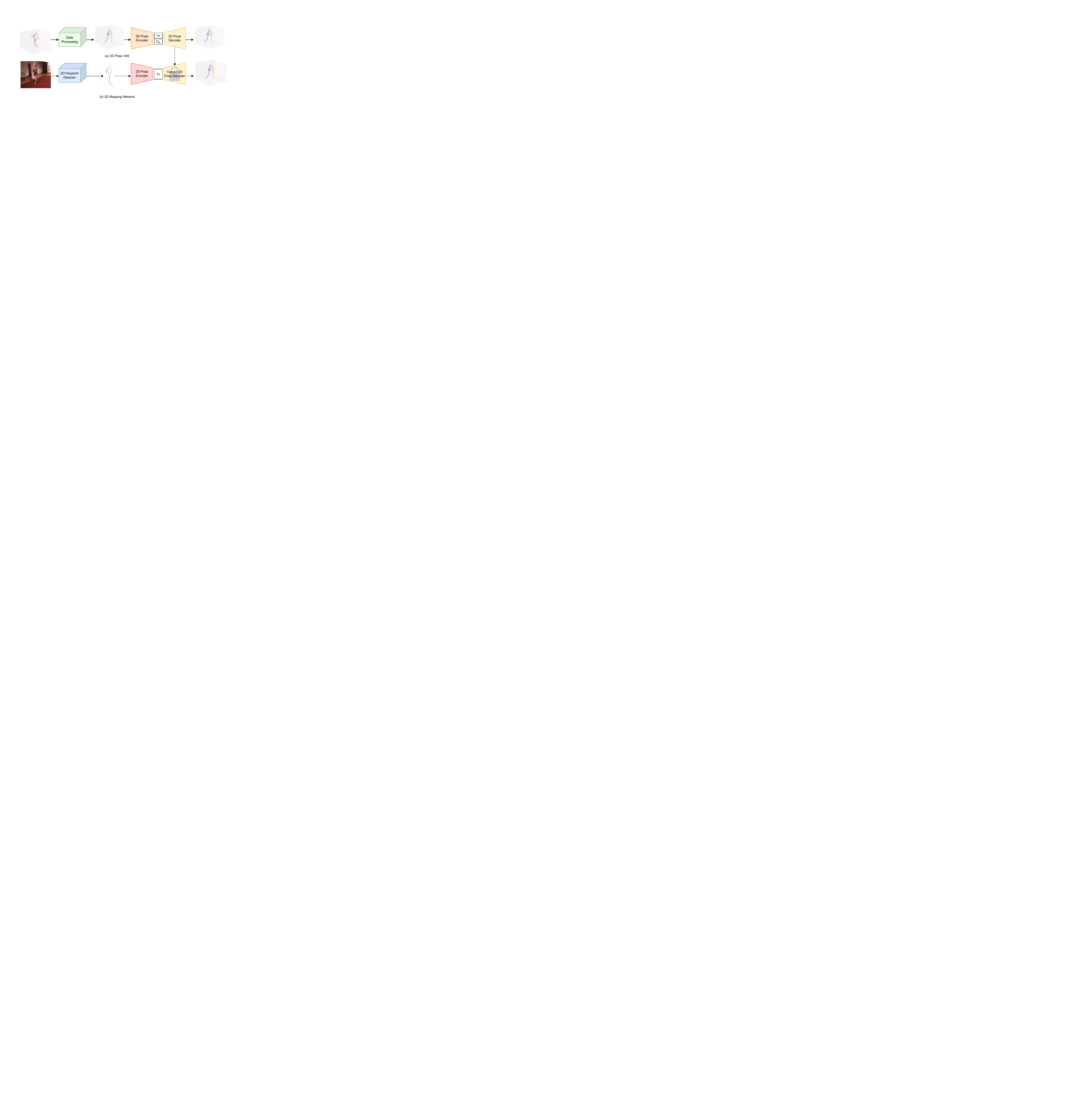}
   \caption{The network on top is our "3D Pose VAE Network." First we pass the 3D input through our data processing phase. Once we have the output we can pass that as input to our VAE network, which generates V-VIPE and then attempts to reconstruct the pose. On the bottom is our "2D Mapping Network." 2D keypoints are extracted using a detector. We then pass these through our 2D encoder and then a locked clone of the decoder network from the 3D Pose VAE Network. This reconstructs the original 3D pose.}
   \label{fig:full_network}
   \vspace{-0.2in}
\end{figure*}

The proposed model consists of two parts, a 3D Pose VAE Network and a 2D Mapping Network. The 3D Pose VAE Network, Figure~\ref{fig:full_network}.a, consists of an encoder network and a decoder network, which make up the VAE model. To stay consistent with other papers we choose ~\cite{DBLP:journals/corr/MartinezHRL17} as the backbone for both our encoder and our decoder.

The benefit of using a VAE for the 3D Pose VAE Network is its ability to generalize to new poses. This is because the goal of a VAE is to synthesize unseen poses. Although this is not our main goal, we do want our network to potentially be able to represent unseen poses, which is a realistic setting in real world applications.

Normalizing the rotation, as defined in the step above, helps the VAE by reducing the range of values that the output can be. We want the VAE to learn all possible human poses within the range and by making that range smaller we make it easier to learn an embedding that spans the whole space. If we omit the rotation realignment then our embedding space would have to learn not only joint location in relation to all other joints, but also joint location in relation to the global space. This is in general unnecessary as location in global space is not relevant when comparing if two poses are equal. Additionally, learning a normalized rotation means that the output is all in one space and can be compared easily without additional alignments.

The 3D Pose VAE Network has two parts: (i) an encoder, which takes as input a 3D pose, $S_{\text{3D}} = \{s_i \in \mathbb{R}^3|i=1 \dots N\}$, where $N$ is the number of keypoints, and outputs a mean for possible embeddings, $\mu_e \in \mathbb{R}^n$, and a variance for the embedding, $\sigma_e \in \mathbb{R}^n$. Using these values and a Gaussian distribution prior we take a sample, $e$. We denote the distribution of the latent space modeled by the encoder with $q(e|S_{\text{3D}})$; (ii) a decoder, which takes in input an embedding, $e$, and outputs an estimation of 3D pose $\hat{S}_{\text{3D}} = \{\hat{s}_i \in \mathbb{R}^3 |i=1 \dots N\}$. The distribution of the decoder is represented as $p(S_{\text{3D}} | e)$.

The goal of the 3D Pose VAE Network is to find a V-VIPE space that is representative of the entire range of 3D human poses for a specific scale and normalization. A feature of the 3D Pose VAE Network should be that poses that are close together in the original 3D space are close together in the embedding space. An important part of learning an accurate mapping from 2D space is that even if there is a slight error in the V-VIPE estimation the output will still be a pose that is similar to the original 3D pose. Additionally, defining a smooth space for V-VIPE enables us to interpret if two poses are close together in 3D space by observing if they are close together in the embedding space.

We define a distance function, $D$, which represents the Mean Per Joint Position Error (MPJPE). MPJPE measures the distance between two 3D points by taking the $L2$ distance between each joint location and then computing the mean of those distances for all joints. 

During training we thus optimize for three factors: 

\begin{itemize}
\item A reconstruction loss, which is equivalent to the Mean Squared Error (MSE) loss between $S_{\text{3D}}$ and $\hat{S}_{\text{3D}}$. 
$L_\text{mse} = \frac{1}{N}\sum(S_{\text{3D}} - \hat{S_{\text{3D}}})^2$
\item The KL Divergence loss
$L_\text{KL} = \text{KL}[q(z|S_{\text{3D}}) | p(z)]$. This loss represents the distance between the distribution of the encoder and the prior distribution, $p(z)$. In this work we use a Gaussian distribution as the prior. 
\item The third is a triplet loss. 
To compute the triplet loss we first find the 3D distances, $D_{i,j}$ within a batch between all elements. For each pose we then set the closest pose in the batch to be the positive example (j) and the second closest pose to be the negative example(k). We make sure the positive and negative poses are at least $.1$ apart from each other and if they aren't we select the next closest pose as the negative example. We do this because we want the examples to be hard, but not too hard that they introduce noise. We compute triplet loss between $i$, $j$ and $k$ by doing $L_\text{triplet} = \max[0, D_{i,k} - D_{i,j} + m]$, where $m$ is our margin. This loss is useful because it causes similar poses to move closer together in the embedding space.

\end{itemize}

This makes the overall loss function to train the 3D Pose VAE:
\begin{equation}
    L_\text{V-VIPE} = L_\text{mse} + L_\text{triplet} + L_\text{KL}
\end{equation}

\subsection{2D Mapping Network}
\label{sec:2d}

Once we have trained the 3D Pose VAE Network we utilize its embedding space to learn a 2D Mapping Network (see Figure~\ref{fig:full_network}.b). In particular, we take the 3D Pose VAE Network decoder model and we freeze it so that it translates from the pre-defined V-VIPE space to 3D coordinates. Next, we train a new encoder $\text{Enc}_{\text{2D}}$ for 2D coordinates. The new encoder takes in input $S_{\text{2D}} = \{p_i \in \mathbb{R}^2|i=1...N\}$ and outputs a V-VIPE, $e \in \mathbb{R}^n$. We pass $e$ through the frozen decoder to get what the embedding represents in 3D space according to the model trained in the previous phase, $\hat{S}_{\text{3D}} = \{p_i \in \mathbb{R}^3|i=1...N\}$.

To train the 2D Mapping Network we use two losses. Given the input, $S_{\text{2D}}$, the output $\hat{S}_{\text{3D}}$ and the ground truth 3D keypoints, $S_{3D}$, we compute $MSE(S_{\text{3D}}, \hat{S}_{\text{3D}})$. We combine this loss with a triplet loss, which we compute similarly as in Section~\ref{sec:vae}. The main difference is that we use the output from the 2D encoder and the ground truth 3D keypoints. We then back-propagate this loss through the whole network, but do not apply the gradient losses to the decoder network. This is because we do not want to change the embedding space, but we just want to train the 2D encoder to make it compliant with the latent space. 

We find that it is beneficial to pre-train the decoder as described in \ref{sec:vae} because we want to construct a space for V-VIPE that is smooth, without also needing to learn a 2D to 3D mapping. Because we train our 3D Pose VAE on normalized 3D poses it will only learn how to map to a normalized pose. Therefore the output of the 2D Mapping Network is also normalized. This means the output is rotation and scale invariant, making it easy to compare 2D poses from different camera viewpoints.

%% file: sec/4_results.tex
\section{Experiments and Results}
\label{sec:exp_res}

\subsection{Experimental Setup}

The model uses a backbone network described described in~\cite{DBLP:journals/corr/MartinezHRL17}. We stack 2 blocks of this network together for both the encoder and the decoder network of both the 3D Pose VAE Network and the 2D Mapping Network. We set the linear size to 1024, and we use a 0.1 dropout. The dimension of a V-VIPE is 32 and the margin for the triplet loss is 1.0. Any 2D keypoint detector could be used, but we chose AlphaPose~\cite{fang2017rmpe, li2019crowdpose, xiu2018poseflow, li2021hybrik}. We use COCO keypoints because they are widely used for 2D detectors. We implemented the model in PyTorch and we trained it on 1 GPU.

\renewcommand{\tabcolsep}{1.2em}
\begin{table*}[t]
\vspace{-0.2in}
\caption{Hit metric results for different values of k. The upper part of the table shows the Hit metrics when using ground truth (GT) keypoints. The bottom part of the table shows the metrics when using keypoint detection(D) and augmentation(A). For Pr-VIPE and our method AlphaPose is the keypoint detector. Epipolar Pose uses its own detector. The $*$ version of Epipolar Pose is trained on the Human3.6 dataset and the $\#$ version is trained on the 3DHP Dataset. Epipolar pose does not generalize to unseen datasets.}
\vspace{0.05in}
\centering
\footnotesize
\begin{tabular} {@{}l | c c c | c c c | c c c @{}}
\toprule
\multicolumn{1}{r|}{Dataset $\rightarrow$} & \multicolumn{3}{c|}{H3.6M} & \multicolumn{3}{c|}{3DHP (All)} & \multicolumn{3}{c}{3DHP (Unseen)}\\ 

\multicolumn{1}{r|}{$k$ $\rightarrow$} & 1 & 10 & 20 & 1 & 10 & 20 & 1 & 10 & 20 \\
\midrule
PR-VIPE (GT) & \textbf{97.6} & \textbf{99.9} & \textbf{100.0} & 42.6 & 72.8 & 79.1 & 43.7 & 73.2 &  82.0\\
Ours (GT) & 89.7 & 98.8 & 99.4 & \textbf{45.3} & \textbf{76.2} & \textbf{83.1} & \textbf{47.9} & \textbf{77.9} & \textbf{84.5} \\
\midrule
2D keypoints & 28.7 & 47.1 & 50.9 & 9.80 & 21.6 & 25.5 & - & - & - \\
$\text{Epipolar Pose}^*$ & 69.0 & 89.7 & 92.7 & - & - & - & - & - & - \\
$\text{Epipolar Pose}^{\#}$ & - & - & - & 24.6 & 53.2 & 61.3 & - & - & - \\
PR-VIPE (D) & \textbf{72.1} & \textbf{94.3} & \textbf{96.8} & 17.9 & 44.7 & 64.1 & 19.2 & 46.6 & 55.6 \\
PR-VIPE (D + A) & 70.9 & 93.1 & 96.0 & 25.4 & 55.6 & 64.1 & 27.8 & 57.7 & 65.8 \\
Ours (D) & 70.0 & 92.7 & 95.6 & 23.5 & 54.3 & 64.0 & 26.2 & 57.0 & 66.4 \\
Ours (D + A) & 69.0 & 93.5 & 96.3 & \textbf{26.9} & \textbf{59.0} & \textbf{68.2} & \textbf{30.1} & \textbf{61.6} & \textbf{70.3} \\
\bottomrule
\end{tabular}
\label{table:hit}
\vspace{-0.1in}
\end{table*}

\subsection{Metrics}
We evaluate the model using two metrics. The first is a hit metric, inspired from~\cite{DBLP:journals/corr/abs-1901-10841}, which we use to measure how often we are able to retrieve a pose that is similar to a query pose. Given two normalized keypoints $S_{\text{3D}}^i$ and $S_{\text{3D}}^j$ we first apply a Procrustes alignment~\cite{RePEc:spr:psycho:v:31:y:1966:i:1:p:1-10} between the two to get $A(S_{\text{3D}}^i)$ and $A(S_{\text{3D}}^j)$. Given a dataset with many views we select two camera views. We find all embeddings for the 2D poses from the selected cameras. Then, we query each embedding from camera 1 and find the $k$ nearest neighbors from the set of embeddings for camera 2. We consider a pair of embeddings a hit if their original 3D pose satisfies $\text{MPJPE}(A(S_{\text{3D}}^i), A(S_{\text{3D}}^j)) < .1$. We report Hit@k for $k$=1,10,20 and average over all pairs of cameras. This metric represents view invariance because it shows how well we can match poses from one viewpoint to similar poses from another viewpoint.

The second is the Mean Per Joint Position Error (MPJPE), which we define in Section~\ref{sec:vae}. This error is used to determine the distance between two sets of 3D keypoints.

\subsection{Datasets}
In all the experiments we train on the standard training set of the Human3.6M dataset (H3.6M)~\cite{h36m_pami}. For our hit metric we use the test set of H3.6M as the validation set and show results on the MPI-INF-3DHP dataset~\cite{mono-3dhp2017} (3DHP).

\noindent
\textbf{Human3.6M}. The H3.6M dataset \cite{h36m_pami} contains 3.6 million human poses taken from 4 different cameras. All of these cameras are at chest level. The standard training set for this dataset is made up of subjects 1,5,6,7 and 8. The standard test set contains poses from subjects 9 and 11. For the evaluation of the hit metric, we follow the method described in~\cite{DBLP:journals/corr/abs-1912-01001}, where they remove poses that are similar. 

\noindent
\textbf{MPI-INF-3DHP}. 3DHP~\cite{mono-3dhp2017} contains 14 different camera angles. For our tasks we remove the overhead cameras, which leaves us with 11 cameras. Of these cameras, 5 are at chest height and the others have a slight vertical angle. This dataset is used to show whether or not our method will generalize to data that is different from the training data.

\subsection{Augmentation}
In order to improve the model's ability to generalize we introduce camera augmentation similar to the work done in~\cite{DBLP:journals/corr/abs-1912-01001}. To calculate this augmentation we take the ground truth 3D pose and randomly rotate it. We then project this pose into 2D. We add augmented poses to each of our batches during training time. We found that it was best to add augmented poses for half of the poses in each batch.

\subsection{Quantitative Results}

\noindent\textbf{Similar Pose Retrieval Experiments.} We compare our model for hit metrics against 3 baselines. The first baseline is the PR-VIPE model, which attempts to define an embedding space without reconstructing the 3D pose; we adopted their open source code and re-trained their model so we would have results on the same 2D pose detector, i.e., AlphaPose. 
The second baseline is simply finding the nearest neighbor of the detected 2D keypoints. The third baseline uses Epipolar Pose~\cite{DBLP:journals/corr/abs-1903-02330} to detect 3D keypoints. In this case, Procrustes alignment is performed between all poses and the closest aligned pose is selected as the match.

We show the hit metrics for the different $k$ values in Table~\ref{table:hit}. The top section of the table shows the results of our method and of PR-VIPE when trained and tested with ground truth (GT) 3D keypoints. The left part of the table reports the results on the test set of H3.6M. We can see that our approach is slightly worse than the PR-VIPE approach. This is because we are testing on very similar data to the original training set. Our model, however, is designed to generalize.
The generalization of the model is demonstrated in the middle part of the table, where we report the performance on the 3DHP dataset when considering all available cameras. In this case, our model gets higher values for all values of $k$. Moreover, when we pair one chest camera with a camera that is not at chest height, i.e., unseen cameras with respect to the training data(right part of the table), we can see that the gap is even larger. For example, when considering $k=$ 1, the gap between the two models is about 4.2 percent for unseen cameras and 2.7 percent for all cameras. This demonstrates that the latent space we acquired during the VAE training is able to generalize to unseen camera viewpoints better than existing models.

\begin{figure*}[t]
  \centering
  \vspace{-.4in}
   \includegraphics[width=\linewidth]{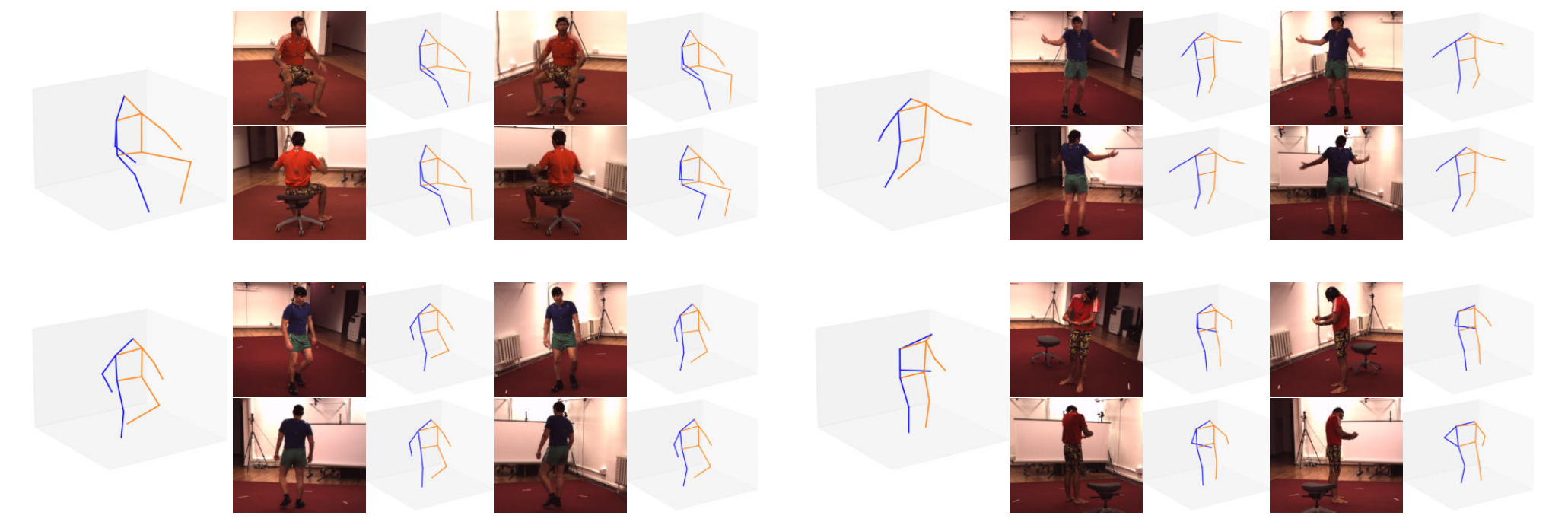}
   \caption{Pose Estimation from 2D images of our model applied to different camera viewpoints. We show 4 sets of results. The ground truth is on the left hand side of each example, while on the right we provide the 4 original views as well as our model 3D output for each view.}
   \vspace{-0.2in}
   
   \label{fig:estimations}
\end{figure*}

\begin{figure*}[htp]
  \centering
   \includegraphics[width=1.0\linewidth]{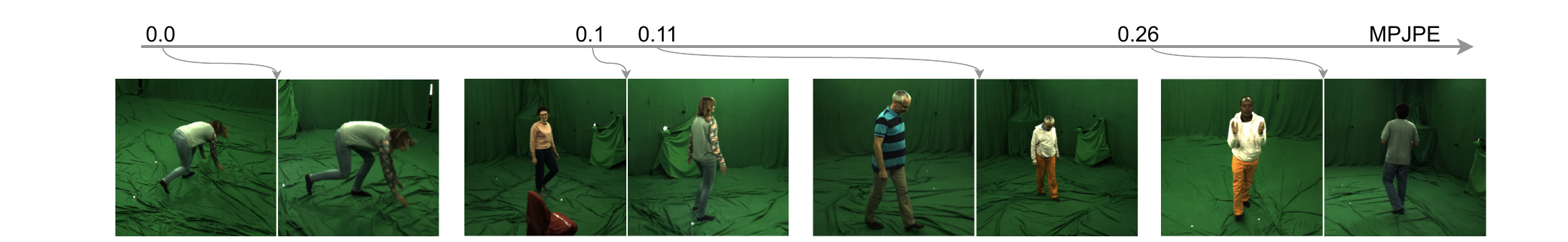}
   \caption{This figure demonstrates what a query and retrieval look like. On the left of each pair of images is the query pose and the image on the right is the image that is considered the closest match by our model. Each pair of images is labeled with the MPJPE between the two poses. Its easy to see that some poses, such as the one on the far left, are easy to retrieve because they are so distinct. And others, such as the one on the far right, have occluded points as well as other factors that make the nearest neighbor hard to find.}
   \vspace{-0.1in}
   \label{fig:retrive}
\end{figure*}

In the bottom section of the table, we show results when the keypoints are automatically detected(D). For PR-VIPE and our model we use AlphaPose. Epipolar Pose detects its own keypoints. Again our method outperforms the PR-VIPE model when generalizing to data different from the training set, 3DHP, as well as to unseen cameras. For example, when $k=$ 10 our method outperforms PR-VIPE by about 5.6 percent for all 3DHP cameras, and by about 7 percent for the unseen category.

In this section we also show results for detected keypoints plus additional training data generated by augmenting the 3D poses. We see an increase from just our detection model for 3DHP because we have introduced new camera viewpoints to the training data. We see an improvement over PR-VIPE when they use augmented data, although we do not get as much of a boost from augmentation because our model already generalizes better than theirs. For $k=$ 1 our model outperforms theirs by 1.5 percent.

Additionally, in the table we report the 2D keypoints and Epipolar Pose results. We can observe that using the 2D keypoints is not effective, as demonstrated by the low hit metric for all $k$ values. The Epipolar Pos$\text{e}^{\#}$ method performs better than both our method and the PR-VIPE method before any augmentation is applied to the data because it is trained on the 3DHP dataset and does not need to generalize. When you try to run the Epipolar Pos$\text{e}^{*}$ model on 3DHP data the output does not resemble human pose. We do not report generalized results for Epipolar pose because of this. Despite the fact that Epipolar Pos$\text{e}^{\#}$ is trained specifically for detection on the 3DHP dataset when we add augmentation of the data to our model we are able to beat their results by about 2 percent. 

\noindent\textbf{3D Pose Estimation Experiments.} In addition to calculating the hit metric described above our model also outputs the predicted 3D pose. We find that the average error of this model is 62.1 millimeters. We calculated this number using a model trained on keypoints detected by the Cascaded Pyramid Network~\cite{DBLP:journals/corr/abs-1711-07319} as this is commonly~\cite{DBLP:journals/corr/PavlakosZDD16, DBLP:journals/corr/abs-1710-06513, DBLP:journals/corr/abs-1901-10841} used for 3D Pose Estimation. We find that while this number is not competitive with current methods for pose estimation that use more complex models or take in more information, such as sequences, it is similar to the error found in \cite{DBLP:journals/corr/MartinezHRL17}, which we use as the backbone for our network.

\subsection{Qualitative Results}

\noindent\textbf{2D to 3D Pose Estimation.} Figure~\ref{fig:estimations} shows examples of our 3D estimations given a 2D image as input. We show examples of 4 different poses each with 4 different camera angles. In the two examples on the left we have very accurate retrievals. All of the cameras have similar retrievals that allow us to determine that the person is in the same pose despite the very different original camera angles. The examples on the right are the ones where our model struggles to find the whole pose. In the example on the top we are able to find the hand position because the hands are visible in every image, however our model struggles to detect that the body is slightly angled. This is likely because the difference in 2D keypoints between an angled and not angled body are very small and our 2D keypoint detector is not accurate enough. In the example on the bottom our model succeeds with the arms, except for one camera viewpoint where the arm is not visible in the image at all. The other way our model struggles is with the head tilt. This is likely because this is difficult to visualize from most camera angles.\looseness=-1

\noindent\textbf{3D Pose Retrieval.} We show how our model is able to retrieve similar poses from different view points. In Figure \ref{fig:retrive} you can see the query pose as well as the pose that is retrieved from a different view point. Ideally, the two poses will be identical. This is the visualization of what the Hit metric represents. If the queried pose is sufficiently close to the retrieved pose then we have a hit.

\begin{figure*}[t]
  \centering
   \includegraphics[width=0.84\linewidth]{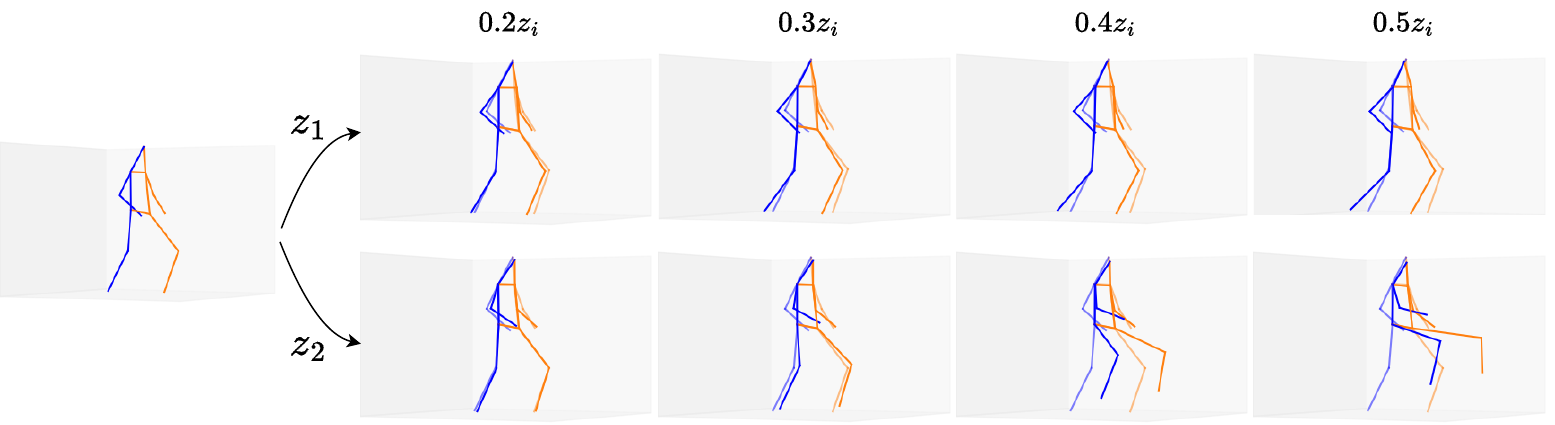}
   \caption{Pose generation, starting from a 3D pose we select two random noise directions $z_i$ and generate poses using increasing magnitudes of noise $\alpha z_i$, where $\alpha\in\{0.2,0.3,0.4,0.5\}$. V-VIPE leads to smooth pose variations and can be used to generate unseen 3D poses.}
   \label{fig:generation}
\end{figure*}

\begin{figure}
\centering
\includegraphics[trim={ 8cm 91cm 86cm 3.25cm}, clip, width=0.9\linewidth]{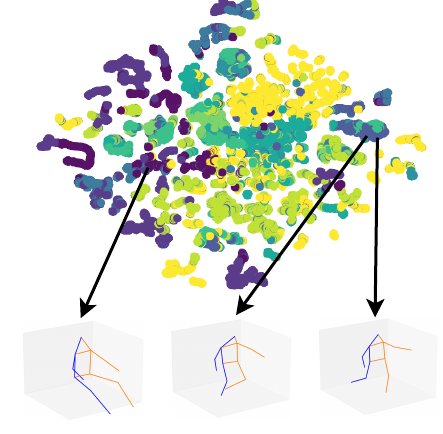}
   \caption{t-SNE visualization of the V-VIPE space of our model for poses in the H3.6M dataset. Each color represents similarity to one of 10 ``key'' poses that we selected. In the expansion, three different poses and their place in the visualization are shown.}
   \label{fig:tsne}
   \vspace{-.15in}
\end{figure}

\noindent\textbf{Visualizing V-VIPE.} Figure~\ref{fig:tsne} shows a t-SNE visualization, which we use to show the smoothness of the learned V-VIPE space, where each dot represents a V-VIPE. In order to properly show the clustering we select 10 visually different 3D poses and color our visualization based on which of the 10 poses is the most similar to the pose that each point represents. It is easy to see from this graph that similar colors are typically found in clusters. This means that the space well represents the notion of similarity between poses. We can see this even clearer in the expansion of the visualization where we show three poses and their locations in the cluster. The two poses on the right are colored the same and are very close together. These are slightly different, but the overall pose is very similar. We then select a point that is very far away and here we can see that the pose is quite different.\looseness=-1

\noindent\textbf{3D Pose Generation.} Our model is able to generate new poses by adding noise to the embedding space of an existing pose. In Figure~\ref{fig:generation} we define a noise array $z$ and add it to an embedding with increasing magnitudes. The pose continues to move in one direction as we increase magnitude showing that our embedding space is smooth.

\section{Ablation Study}
\label{sec:ablation}
We performed an ablative analysis in order to understand which of our design choices best contributed to our results.
\noindent\textbf{Triplet Loss.}
First we examine how important it is that we include the triplet loss term in our method. We remove it from the loss term and find that the new Hit@1 value is 17.41 with no augmented data. This is a drop of 6.1 from the Hit@1 value when triplet loss is included. Therefore the triplet loss value is important to the overall loss term.

\noindent\textbf{Data Processing.}
We examine how important it is for us to rotate the 3D pose before training on our model. This step is important because it enables us to compare the similarity of poses with two different global rotations without needing to do a time consuming Procrustes Alignment between every pair of poses. We find that the Hit@1 value on 3DHP with no augmentation obtained when using non rotated points is 18.0 percent, a 5.5 percent decline from our approach.

\noindent\textbf{Pretraining the Decoder.}
Finally, we studied whether or not pretraining a VAE and using a defined embedding space contributed to our final hit metric. We found that the Hit@1 value for the model with no pretraining is 23.4 versus the 23.5 we obtained by completing the pretraining step. However, this step is important anyways because it enables the model to do 3D Pose Retrieval. Without it we would not be able to map our 3D poses to our embedding space. Therefore we would not be able to generate similar poses to a given 3D pose or query a 3D pose to find a similar 2D pose from a set of images.

%% file: sec/5_conclusion.tex
\section{Conclusion}
\label{sec:conclusion}
In this work we showed that by using only 3D poses to define a V-VIPE space we can define a better camera invariant space than if we were to only use 2D poses. We defined a procedure made of two steps: first we train a VAE model to learn a latent space of 3D poses; then, we train a 2D keypoints encoder that is linked to the VAE decoder to allow 3D reconstructions of 2D images. We adopted a VAE model as it creates a smooth latent space that can generalize better towards unseen poses during training. In order to achieve this goal, we train a VAE with a three component loss function. We performed an extensive experimental evaluation, by using two datasets, i.e., Human3.6M and MPI-INF-3DHP. We demonstrated that the latent space is modeling a meaningful notion of similarity of the embeddings. This is reflected in the Pose Retrieval experiments where we improve about 2.5 percent in the Hit@1 metric when considering unseen cameras. We also showed qualitative examples demonstrating the capability of our embedding space to capture the notion of similarity of poses. This is important in downstream tasks. In the future we believe that this approach has a lot of promise for application to downstream tasks such as action segmentation and detection. \\
\noindent\textbf{Acknowledgements:} This work was partially supported by NSF CAREER Award (\#2238769) to AS and the DARPA SAIL-ON (W911NF2020009) program.

%% file: sec/X_suppl.tex
\section{Contents}
Included in the supplementary material are several extra pages of results. In Figure~\ref{fig:noisey} we include several results that show the effect of adding noise to an embedding for a pose. In Figure~\ref{fig:waypoints} we show that by finding embeddings in between two poses we can generate way point poses. Figure~\ref{fig:retrievals} shows the results of querying several images. 


\begin{figure*}
  \centering
   \includegraphics[width=\linewidth]{Noisy.drawio (1).pdf}
   \caption{In this figure we show what happens when we add random noise to the embedding space for a pose. For each original pose we sample noise and then add it to the original embedding in larger magnitudes. The smallest magnitude is on the left and we increase the magnitude of noise as the images move to the right.}
   \label{fig:noisey}
\end{figure*}

\begin{figure*}
  \centering
   \includegraphics[width=\linewidth]{In Between.drawio (1).pdf}
   \caption{Given the frame on the left and the frame on the right for each example we show our models ability to generate frames in between two poses. We take the mean for the embedding space of two poses, calculate the distance between those two means and then add a portion of the distance to the embedding on the left at each step. The images seen in the middle three columns are all generated. We can see that we are able to generate a sequence of poses that lead from one random pose to another.}
   \label{fig:waypoints}
\end{figure*}


\begin{figure*}
  \centering
  \includegraphics[width=\linewidth]{Retrievals.drawio (4) (1).pdf}
  \caption{Several retrievals from the 3DHP dataset. On the left is the query pose and on the right is the retrieved pose. On top of each pair of images is the distance between the two poses. }
  \label{fig:retrievals}
\end{figure*}

%% file: main.bbl
\begin{thebibliography}{32}
\providecommand{\natexlab}[1]{#1}
\providecommand{\url}[1]{\texttt{#1}}
\expandafter\ifx\csname urlstyle\endcsname\relax
  \providecommand{\doi}[1]{doi: #1}\else
  \providecommand{\doi}{doi: \begingroup \urlstyle{rm}\Url}\fi

\bibitem[Chen et~al.(2019)Chen, Tyagi, Agrawal, Drover, Rohith, Stojanov, and Rehg]{DBLP:journals/corr/abs-1904-04812}
Ching{-}Hang Chen, Ambrish Tyagi, Amit Agrawal, Dylan Drover, M.~V. Rohith, Stefan Stojanov, and James~M. Rehg.
\newblock Unsupervised 3d pose estimation with geometric self-supervision.
\newblock \emph{CoRR}, abs/1904.04812, 2019.

\bibitem[Chen et~al.(2017)Chen, Wang, Peng, Zhang, Yu, and Sun]{DBLP:journals/corr/abs-1711-07319}
Yilun Chen, Zhicheng Wang, Yuxiang Peng, Zhiqiang Zhang, Gang Yu, and Jian Sun.
\newblock Cascaded pyramid network for multi-person pose estimation.
\newblock \emph{CoRR}, abs/1711.07319, 2017.

\bibitem[Cheng et~al.(2020)Cheng, Yang, Wang, and Tan]{DBLP:journals/corr/abs-2004-11822}
Yu Cheng, Bo Yang, Bo Wang, and Robby~T. Tan.
\newblock 3d human pose estimation using spatio-temporal networks with explicit occlusion training.
\newblock \emph{CoRR}, abs/2004.11822, 2020.

\bibitem[Fang et~al.(2017{\natexlab{a}})Fang, Xu, Wang, Liu, and Zhu]{DBLP:journals/corr/abs-1710-06513}
Haoshu Fang, Yuanlu Xu, Wenguan Wang, Xiaobai Liu, and Song{-}Chun Zhu.
\newblock Learning knowledge-guided pose grammar machine for 3d human pose estimation.
\newblock \emph{CoRR}, abs/1710.06513, 2017{\natexlab{a}}.

\bibitem[Fang et~al.(2017{\natexlab{b}})Fang, Xie, Tai, and Lu]{fang2017rmpe}
Hao-Shu Fang, Shuqin Xie, Yu-Wing Tai, and Cewu Lu.
\newblock {RMPE}: Regional multi-person pose estimation.
\newblock In \emph{ICCV}, 2017{\natexlab{b}}.

\bibitem[Girdhar et~al.(2016)Girdhar, Fouhey, Rodriguez, and Gupta]{DBLP:journals/corr/GirdharFRG16}
Rohit Girdhar, David~F. Fouhey, Mikel Rodriguez, and Abhinav Gupta.
\newblock Learning a predictable and generative vector representation for objects.
\newblock \emph{CoRR}, abs/1603.08637, 2016.

\bibitem[Hossain and Little(2017)]{DBLP:journals/corr/abs-1711-08585}
Mir Rayat~Imtiaz Hossain and James~J. Little.
\newblock Exploiting temporal information for 3d pose estimation.
\newblock \emph{CoRR}, abs/1711.08585, 2017.

\bibitem[Ionescu et~al.(2014)Ionescu, Papava, Olaru, and Sminchisescu]{h36m_pami}
Catalin Ionescu, Dragos Papava, Vlad Olaru, and Cristian Sminchisescu.
\newblock Human3.6m: Large scale datasets and predictive methods for 3d human sensing in natural environments.
\newblock \emph{IEEE Transactions on Pattern Analysis and Machine Intelligence}, 2014.

\bibitem[Kabsch(1978)]{Kabsch:a15629}
W. Kabsch.
\newblock {A discussion of the solution for the best rotation to relate two sets of vectors}.
\newblock \emph{Acta Crystallographica Section A}, 34\penalty0 (5):\penalty0 827--828, 1978.

\bibitem[Katircioglu et~al.(2018)Katircioglu, Tekin, Salzmann, Lepetit, and Fua]{Katircioglu2018LearningLR}
Isinsu Katircioglu, Bugra Tekin, Mathieu Salzmann, Vincent Lepetit, and Pascal~V. Fua.
\newblock Learning latent representations of 3d human pose with deep neural networks.
\newblock \emph{International Journal of Computer Vision}, 126:\penalty0 1326--1341, 2018.

\bibitem[Kocabas et~al.(2019)Kocabas, Karagoz, and Akbas]{DBLP:journals/corr/abs-1903-02330}
Muhammed Kocabas, Salih Karagoz, and Emre Akbas.
\newblock Self-supervised learning of 3d human pose using multi-view geometry.
\newblock \emph{CoRR}, abs/1903.02330, 2019.

\bibitem[Li et~al.(2019)Li, Wang, Zhu, Mao, Fang, and Lu]{li2019crowdpose}
Jiefeng Li, Can Wang, Hao Zhu, Yihuan Mao, Hao-Shu Fang, and Cewu Lu.
\newblock Crowdpose: Efficient crowded scenes pose estimation and a new benchmark.
\newblock In \emph{Proceedings of the IEEE/CVF conference on computer vision and pattern recognition}, pages 10863--10872, 2019.

\bibitem[Li et~al.(2021)Li, Xu, Chen, Bian, Yang, and Lu]{li2021hybrik}
Jiefeng Li, Chao Xu, Zhicun Chen, Siyuan Bian, Lixin Yang, and Cewu Lu.
\newblock Hybrik: A hybrid analytical-neural inverse kinematics solution for 3d human pose and shape estimation.
\newblock In \emph{Proceedings of the IEEE/CVF Conference on Computer Vision and Pattern Recognition}, pages 3383--3393, 2021.

\bibitem[Ma et~al.(2021)Ma, Chen, Kong, Wang, Liu, Tang, Yan, Xie, Lin, and Xie]{DBLP:journals/corr/abs-2110-09554}
Haoyu Ma, Liangjian Chen, Deying Kong, Zhe Wang, Xingwei Liu, Hao Tang, Xiangyi Yan, Yusheng Xie, Shih{-}Yao Lin, and Xiaohui Xie.
\newblock Transfusion: Cross-view fusion with transformer for 3d human pose estimation.
\newblock \emph{CoRR}, abs/2110.09554, 2021.

\bibitem[Martinez et~al.(2017)Martinez, Hossain, Romero, and Little]{DBLP:journals/corr/MartinezHRL17}
Julieta Martinez, Rayat Hossain, Javier Romero, and James~J. Little.
\newblock A simple yet effective baseline for 3d human pose estimation.
\newblock \emph{CoRR}, abs/1705.03098, 2017.

\bibitem[Mehta et~al.(2017)Mehta, Rhodin, Casas, Fua, Sotnychenko, Xu, and Theobalt]{mono-3dhp2017}
Dushyant Mehta, Helge Rhodin, Dan Casas, Pascal Fua, Oleksandr Sotnychenko, Weipeng Xu, and Christian Theobalt.
\newblock Monocular 3d human pose estimation in the wild using improved cnn supervision.
\newblock In \emph{3D Vision (3DV), 2017 Fifth International Conference on}. IEEE, 2017.

\bibitem[Panda and Mukherjee(2021)]{9506722}
Aditya Panda and Dipti~Prasad Mukherjee.
\newblock Monocular 3d human pose estimation by multiple hypothesis prediction and joint angle supervision.
\newblock In \emph{2021 IEEE International Conference on Image Processing (ICIP)}, pages 3243--3247, 2021.

\bibitem[Park et~al.(2016)Park, Hwang, and Kwak]{DBLP:journals/corr/ParkHK16}
Sungheon Park, Jihye Hwang, and Nojun Kwak.
\newblock 3d human pose estimation using convolutional neural networks with 2d pose information.
\newblock \emph{CoRR}, abs/1608.03075, 2016.

\bibitem[Pavlakos et~al.(2016)Pavlakos, Zhou, Derpanis, and Daniilidis]{DBLP:journals/corr/PavlakosZDD16}
Georgios Pavlakos, Xiaowei Zhou, Konstantinos~G. Derpanis, and Kostas Daniilidis.
\newblock Coarse-to-fine volumetric prediction for single-image 3d human pose.
\newblock \emph{CoRR}, abs/1611.07828, 2016.

\bibitem[Remelli et~al.(2020)Remelli, Han, Honari, Fua, and Wang]{DBLP:journals/corr/abs-2004-02186}
Edoardo Remelli, Shangchen Han, Sina Honari, Pascal Fua, and Robert Wang.
\newblock Lightweight multi-view 3d pose estimation through camera-disentangled representation.
\newblock \emph{CoRR}, abs/2004.02186, 2020.

\bibitem[Schönemann(1966)]{RePEc:spr:psycho:v:31:y:1966:i:1:p:1-10}
Peter Schönemann.
\newblock A generalized solution of the orthogonal procrustes problem.
\newblock \emph{Psychometrika}, 31\penalty0 (1):\penalty0 1--10, 1966.

\bibitem[Sharma et~al.(2019)Sharma, Varigonda, Bindal, Sharma, and Jain]{DBLP:journals/corr/abs-1904-01324}
Saurabh Sharma, Pavan~Teja Varigonda, Prashast Bindal, Abhishek Sharma, and Arjun Jain.
\newblock Monocular 3d human pose estimation by generation and ordinal ranking.
\newblock \emph{CoRR}, abs/1904.01324, 2019.

\bibitem[Sun et~al.(2019)Sun, Zhao, Chen, Schroff, Adam, and Liu]{DBLP:journals/corr/abs-1912-01001}
Jennifer~J. Sun, Jiaping Zhao, Liang{-}Chieh Chen, Florian Schroff, Hartwig Adam, and Ting Liu.
\newblock View-invariant probabilistic embedding for human pose.
\newblock \emph{CoRR}, abs/1912.01001, 2019.

\bibitem[Sun et~al.(2017)Sun, Xiao, Liang, and Wei]{DBLP:journals/corr/abs-1711-08229}
Xiao Sun, Bin Xiao, Shuang Liang, and Yichen Wei.
\newblock Integral human pose regression.
\newblock \emph{CoRR}, abs/1711.08229, 2017.

\bibitem[Wang et~al.(2021)Wang, Zhang, Cai, Yan, and Feng]{DBLP:journals/corr/abs-2111-04076}
Tao Wang, Jianfeng Zhang, Yujun Cai, Shuicheng Yan, and Jiashi Feng.
\newblock Direct multi-view multi-person 3d pose estimation.
\newblock \emph{CoRR}, abs/2111.04076, 2021.

\bibitem[Wei et~al.(2019)Wei, Lan, Zeng, and Chen]{DBLP:journals/corr/abs-1901-10841}
Guoqiang Wei, Cuiling Lan, Wenjun Zeng, and Zhibo Chen.
\newblock View invariant 3d human pose estimation.
\newblock \emph{CoRR}, abs/1901.10841, 2019.

\bibitem[Xia and Xiao(2020)]{9257470}
Hailun Xia and Meng Xiao.
\newblock 3d human pose estimation with generative adversarial networks.
\newblock \emph{IEEE Access}, 8:\penalty0 206198--206206, 2020.

\bibitem[Xiu et~al.(2018)Xiu, Li, Wang, Fang, and Lu]{xiu2018poseflow}
Yuliang Xiu, Jiefeng Li, Haoyu Wang, Yinghong Fang, and Cewu Lu.
\newblock {Pose Flow}: Efficient online pose tracking.
\newblock In \emph{BMVC}, 2018.

\bibitem[Yang et~al.(2018)Yang, Ouyang, Wang, Ren, Li, and Wang]{DBLP:journals/corr/abs-1803-09722}
Wei Yang, Wanli Ouyang, Xiaolong Wang, Jimmy S.~J. Ren, Hongsheng Li, and Xiaogang Wang.
\newblock 3d human pose estimation in the wild by adversarial learning.
\newblock \emph{CoRR}, abs/1803.09722, 2018.

\bibitem[Zeng et~al.(2020)Zeng, Sun, Huang, Liu, Xu, and Lin]{DBLP:journals/corr/abs-2007-09389}
Ailing Zeng, Xiao Sun, Fuyang Huang, Minhao Liu, Qiang Xu, and Stephen Lin.
\newblock Srnet: Improving generalization in 3d human pose estimation with a split-and-recombine approach.
\newblock \emph{CoRR}, abs/2007.09389, 2020.

\bibitem[Zheng et~al.(2021)Zheng, Zhu, Mendieta, Yang, Chen, and Ding]{DBLP:journals/corr/abs-2103-10455}
Ce Zheng, Sijie Zhu, Mat{\'{\i}}as Mendieta, Taojiannan Yang, Chen Chen, and Zhengming Ding.
\newblock 3d human pose estimation with spatial and temporal transformers.
\newblock \emph{CoRR}, abs/2103.10455, 2021.

\bibitem[Zhou et~al.(2017)Zhou, Huang, Sun, Xue, and Wei]{DBLP:journals/corr/ZhouH0XW17}
Xingyi Zhou, Qixing Huang, Xiao Sun, Xiangyang Xue, and Yichen Wei.
\newblock Weakly-supervised transfer for 3d human pose estimation in the wild.
\newblock \emph{CoRR}, abs/1704.02447, 2017.

\end{thebibliography}
